\documentclass{bmvc2k}

\usepackage{times}
\usepackage{epsfig}
\usepackage{graphicx}
\usepackage{amsmath}
\usepackage{amssymb}
\usepackage{mathtools}
\usepackage{multirow}
\usepackage{booktabs}

\title{Adaptive Compression-based Lifelong Learning}

\addauthor{Shivangi Srivastava}{shivangi.srivastava@wur.nl}{1}
\addauthor{Maxim Berman}{maxim.berman@kuleuven.be}{2}
\addauthor{Matthew B.\ Blaschko}{matthew.blaschko@esat.kuleuven.be}{2}
\addauthor{Devis Tuia}{devis.tuia@wur.nl}{1}

\addinstitution{
 Geo-Information Science and Remote Sensing, Wageningen University\\ the Netherlands
}
\addinstitution{
 Center for Processing Speech \& Images, KU Leuven\\ Belgium
}

\runninghead{Srivastava et al.}{Adaptive Compression-based Lifelong Learning}

\def\st{\emph{s.t}\bmvaOneDot}
\def\ie{\emph{i.e}\bmvaOneDot}
\def\eg{\emph{e.g}\bmvaOneDot}

\begin{document}

\maketitle

\begin{abstract}
The problem of a deep learning model losing performance on a previously learned task when fine-tuned to a new one is a phenomenon known as \emph{Catastrophic forgetting}. There are two major ways to mitigate this problem: either preserving activations of the initial network during training with a new task; or restricting the new network activations to remain close to the initial ones. 
The latter approach falls under the denomination of \emph{lifelong learning}, where the model is updated in a way that it performs well on both old and new tasks, without having access to the old task's training samples anymore.

Recently, approaches like pruning networks for freeing network capacity during sequential learning of tasks have been gaining in popularity. Such approaches allow learning small networks while making redundant parameters available for the next tasks. The common problem encountered with these approaches is that the pruning percentage is hard-coded, irrespective of the number of samples, of the complexity of the learning task and of the number of classes in the dataset. We propose a method based on Bayesian optimization to perform adaptive compression/pruning of the network and show its effectiveness in lifelong learning. Our method learns to perform heavy pruning for small and/or simple datasets while using milder compression rates for large and/or complex data. Experiments on classification and semantic segmentation demonstrate the applicability of learning network compression, where we are able to effectively preserve performances along sequences of tasks of varying complexity.
\end{abstract}

\section{Introduction}
Humans are very good at learning tasks in a sequence~\cite{cichon2015branch}, including the case when observations from the previous tasks are not accessible anymore. On the contrary, artificial intelligence-based learning models, such as Convolutional Neural Networks (CNNs), struggle in that situation: when confronted with the new task, CNNs tend to migrate towards it and to forget the representation that helped to solve the original task. This problem is generally known as \emph{catastrophic forgetting}~\cite{mccloskey1989catastrophic, ratcliff1990connectionist, mcclelland1995there, french1999catastrophic, kumaran2016learning}. After some initial empirical attempts to understand the phenomenon~\cite{srivastava2013compete, goodfellow2013empirical}, methods dealing explicitly with the problem have been proposed in the literature under the name of \emph{lifelong learning}~\cite{li2018learning,rannen2017encoder}: those methods usually either preserve activations of the initial network when training for new task, or constrain the new network's activations to remain close to the initial ones. 
The promise of lifelong learning is to provide methods that are able to perform well on both tasks, even after having learned them in a sequence and without access to the labels of the former task while learning for the next one. 
 
In parallel, approaches known as \emph{pruning networks} for freeing network capacity during sequential learning of tasks have been gaining in popularity~\cite{mallya2018packnet}. The weights in the network associated with each task are pruned until they occupy a fraction of the global network capacity; these pruned weights then remains frozen while learning the subsequent tasks. 
By doing so, one can provide capacity for learning the new tasks without having to significantly increase the model size. Moreover, such a strategy also allows reusing redundant parameters for the next tasks, while restricting the growth of the model only to a new classifier layer per new task being considered.

Pruning networks primarily rely on one parameter, the pruning percentage, which balances the compression gain and accuracy decrease. Even in the most recent models, this percentage is generally treated as a hyperparameter, and hence hard-coded. However, we argue that hard-coding the percentage is suboptimal, due to multiple reasons. First, the compression rate needs to be related to the size and complexity of the task at hand: while it makes sense to prune heavily, \ie heavier network compression rates, for a small and/or simple dataset, it is more advisable to perform lower compression for larger or more complex datasets (like ImageNet). Second, the order in which the tasks are coming is also of importance: one cannot know in advance when the more complex task will come, so the ability to save as much capacity as possible is a desirable property for a sequential learning algorithm.

In this paper, we tackle the problem of learning compression of neural networks for sequential learning. Using a Bayesian optimization approach, we learn the optimal compression rate to be applied, which is optimal in the sense that it will perform compression up to an acceptable loss in performance in the previous task. By doing so, the method \emph{guarantees} to avoid catastrophic forgetting, while saving as much network capacity as possible for the next task(s). Additionally, and since the weights on the previous tasks are never modified, there is no need to actively train for the preservation of the accuracy on the previous tasks. We showcase the interest of learning compressing CNNs both in image classification and segmentation: in the first case, we show how a learned compression rate can save capacity to learn a complex new task like ImageNet, while in the second we showcase the advantage of our proposed method in a three-tasks satellite image segmentation problem.

\section{Related Works}\label{sec:related_works}

The most common way to learn a new task from a model trained on another is to fine-tune it~\cite{girshick2014rich,donahue2014decaf}. Fine-tuning works generally very well for the new task, but at the price of a drop in accuracy for the former, since the weights are modified and tuned for the new task. A first possible solution is to keep a copy of the original model trained on the original task, but this leads to heavy memory requirements with an increase in the number of tasks. Another solution would be to perform multi-task learning~\cite{caruana1997multitask}, but this strategy relies on labeled data for all tasks to be available during training, which is typically not possible in sequential learning.

The issue of accessing the data of previous tasks is mitigated to a large extent in the `Learning without Forgetting' (LwF) framework \cite{li2016LearningWF, li2018learning}. LwF combines fine-tuning and distillation networks~\cite{hinton2015distilling}, where a knowledge distillation loss~\cite{hinton2015distilling} tries to preserve the output of the former classifier on data from the new task. However, LwF uses several losses, whose number (and balancing weights involved) scales linearly with the number of tasks. The authors in \cite{kirkpatrick2017overcoming,lee2017overcoming} propose approaches where the distance between parameters of the models trained on the old and new tasks is regulated via $\ell_2^2$ losses. As for LwF, the number of parameters increases with the number of tasks. In \cite{rannen2017encoder}, the authors use autoencoders in addition to LwF. This approach has an overhead of a linearly increasing number of autoencoders and task-specific classifiers, several hyperparameters and also a distillation loss between the single-task and the multitask model, making its training complex.

An alternative direction to the above is the idea of removing redundant parameters by neural network compression~\cite{mallya2018packnet}. The authors report good results but only use a fixed pruning percentage for all the tasks, irrespective of the complexity of the data involved. Other works have used masks on networks' weights, either using attention~\cite{serra2018overcatastrforget} or by learning binary weights masks end-to-end, in order to use only the weights useful for the new task~\cite{mallya2018piggyback}.

Fine-tuning or lifelong learning approaches lead to an automatically learned balance between the network capacity dedicated to the old task and the network capacity dedicated to the new task through the learning objective but do not guarantee the preservation of the performance of the network on older tasks. On the contrary, compression-based approaches allow a stronger guarantee no forgetting through the preservation of the former task weights in the network but require a manual and arbitrary adjustment of the network capacity dedicated to older tasks vs. the new task. 
Our proposed framework represents the best of both worlds; given a sequence of tasks, it learns optimal compression rates for each task and avoids drops in accuracy by allocating parts of its tunable parameters to the different tasks in advance. The amount of allocated memory depends on how much compression is applied, and this rate is learned from the data itself through Bayesian optimization.

\section{Adaptive compression-based lifelong learning (AcLL)}\label{sec:our_approach}

The compression-based lifelong learning approach of \cite{mallya2018packnet} prescribes a fixed pruning rate in order to compress a neural network. A model trained for a particular task after pruning frees up parameters that can be used to learn other tasks. The set of weights that are set to zero are stored as a bit mask. In training the next task, the weights that had been previously set to zero are optimized to maximize performance on the new task, disregarding their effect on the previous task, while the weights that have been retained from the first task are fixed in perpetuity. At test time, the bit mask is applied when evaluating samples from the first task to ensure that the performance of the network is unaffected by the weight changes coming from subsequent tasks. This way, catastrophic forgetting is avoided, and performance on earlier tasks is never degraded by training subsequent tasks. However, this comes at a cost: the number of weights that can be modified to train subsequent tasks is reduced. \cite{mallya2018packnet} propose a fixed pruning weight (either 50\% or 75\% of remaining weights), meaning after the first task 50\% of weights remain for the second task, after which 25\% remains for the 3rd task, 12.5\% for the 4th task, and so on. Assuming the $n$th task requires a minimum fixed number of tunable parameters to achieve reasonable accuracy, the original network would need to have a size exponential in $n$ with this fixed weighting scheme. We address this by not setting a compression rate \emph{a priori}, but {\emph{adaptively}}.

Compression algorithms are typically parametrized, \emph{e.g.} through the rank in a low-dimensional matrix factorization, a threshold, or a fraction of weights to remove in sparsifying a network, with each parameter setting achieving a different amount of compression. As a result, there is a trade-off between the amount of network compression and the accuracy of the resulting compressed network. Our intuition is that the amount of compression at any stage of a lifelong learning algorithm should be determined by a performance target on a given task. Then, the amount of compression can be maximized subject to this performance target. Let $f$ be an unmodified neural network, $\theta$ a vector of compression parameters, and $f_\theta$ the resulting compressed network. Also, let $\mathcal{R}(f)$ be the risk of a function $f$. $\mathcal{R}(f)$ is  the 0-1 loss and the models are optimized on a validation set from the current task. We then consider the following optimization problem:
\begin{align}
\min_{\theta\in \mathbb{R}^d} & \operatorname{size}(f_\theta) \label{eq:ConstrainedCompression} \\
\text{\st} \,& \mathcal{R}(f_\theta) \leq \mathcal{R}(f) + \varepsilon
\end{align}
where $\varepsilon$ is typically greater than zero and indicates the amount of loss over an uncompressed network that will be tolerated in order to reserve network capacity for future tasks.\footnote{We have observed that, particularly for small amounts of data, a degree of compression can provide a regularizing effect and it is possible to achieve lower risk from a compressed network than an original uncompressed network.} The tolerance $\varepsilon$ is a user-defined parameter that trades off the acceptable loss in performance on the current task versus higher compression rates (and the corresponding  available model capacity  for new tasks).  In previous work with hand-selected parameters~\cite{mallya2018packnet}, the reduction in accuracy is of the order of 1-2\%.

The optimization of Eq.~\eqref{eq:ConstrainedCompression} is not immediately evident, as the size of the function is not differentiable, and the constraint is over a complicated (non-differentiable in the case of \eg a 0-1 loss) risk functional. 
In the following, we formulate this problem using a Lagrangian-based optimization strategy to transform it into a series of unconstrained optimization problems. 
Subsequently, we solve these unconstrained problems using Bayesian optimization.

The Lagrangian of our constrained optimization is
\begin{align} \label{eq:ConstrainedCompressionLagrangian}
    \mathcal{L}(\theta,\lambda) := \operatorname{size}(f_\theta) + \lambda \left(\mathcal{R}(f_\theta) - (\mathcal{R}(f) + \varepsilon)\right),
\end{align}
which indicates that for varying $\lambda \geq 0$, each optimization over $\theta$ will be of the form
\begin{equation}\label{eq:lagrangian_subproblem}\arg\min_\theta \mathcal{L}(\theta,\lambda)= \arg\min_{\theta} \operatorname{size}(f_\theta) + \lambda\mathcal{R}(f_\theta).
\end{equation}
For fixed $\lambda$, we call an off-the-shelf Bayesian optimization routine~\cite{BayesianOptimizationGithub}; indeed, Bayesian optimization can be considered to be at the state of the art for optimization of black-box, non-differentiable functions \cite{brochu2010,Frazier2018}. 

We wish to determine the optimal $\lambda$ suited for a target accuracy tolerance $\epsilon$ in the original problem~\eqref{eq:ConstrainedCompression}. 
In general, one could solve the unconstrained problem~\eqref{eq:lagrangian_subproblem} many times for different values of $\lambda$. 
This is however inefficient, in particular because of the time spent for the multiple evaluations of $\mathcal{R}(f_\theta)$ needed for the bayesian optimization of problem~\eqref{eq:lagrangian_subproblem}. We use two properties in order to reduce the optimization time:
\begin{itemize}
    \item \textbf{Concavity.} As Eq.~\eqref{eq:lagrangian_subproblem} is concave in $\lambda$ \cite[Sec.~5.1.2]{Boyd:2004:CO:993483}, the search for the optimal $\lambda$ for a given $\epsilon$ can be made more efficient by using line search strategies: this can be done e.g. using a cutting plane approach -- where each evaluation of the dual~\eqref{eq:lagrangian_subproblem} gives a subgradient direction.
    In our experiments, we found that performing a binary search over $\lambda$, which is equivalent to taking the sign of that subgradient direction, already leads to an acceptable convergence of $\lambda$.
    \item {\textbf{Efficient caching strategy.}} We cache each evaluation of $\mathcal{R}(f_\theta)$ and reuse these evaluations among different Bayesian optimization runs, for different values of $\lambda$. 
    Indeed,
    given observations at iteration $t-1$ of the Lagrangian optimization used to model $\operatorname{size}(f_\theta) + \lambda_{t-1}\mathcal{R}(f_\theta)$, it is straightforward to simply re-weight the saved values by a different factor $\lambda_t$ to initialize the Gaussian process model for the next round of Bayesian optimization. 
    This seeding of the Bayesian optimization process leads to a faster convergence and reduces the number of new evaluations needed.
\end{itemize}

In practice, the cost of optimizing the constrained 
{form} with this caching scheme is a very small multiple of the cost of a single unconstrained optimization. In our experiments using pruning-based compression, we have observed an overall increase in the cost of compression by a factor of approximately 6 to 8 using this Lagrangian-based optimization scheme with caching vs.\ a fixed compression ratio.

\section{Experiments}\label{sec:experiments_results}
In this section, we present the results on two challenging settings: sequential learning of models for classification with increasing complexity (Section~\ref{sec:class}) and sequential learning of models for semantic segmentation of satellite images (Section~\ref{sec:ss}).

\subsection{Classification}\label{sec:class}
In this section, we performed experiments to verify if adaptive compression can lead to better classification performances on a complex task. We experimentally show that by applying our adaptive compression method we perform stronger compression to a model trained with a relatively simple task. This helps to improve the performance of the model for a subsequent more complex task that will have more free parameters to train. It is worth mentioning that after we train the model for the second more complex task the model is still able to perform prediction in the first task with accuracy within tolerated limits. Our motivation comes from~\cite{mallya2018packnet}, where the authors first trained a model on ImageNet~\cite{russakovsky2015imagenet}, pruned 50\% of the model weights, but then applied lifelong learning to a smaller CUBS Birds dataset~\cite{wah2011caltech}, for which we argue not a large capacity is necessary, so \emph{a priori} small compression rates are acceptable.

 \paragraph{Data and setup.}
As a first task, we trained a model on the CUBS Birds dataset and then switched to ImageNet as a second task. Details on the number of images are provided in Table\ref{tab:dataclass}. We used ResNet50~\cite{he2016deep} as base model, initialized with the pretrained weights of Places365~\cite{zhou2018places}. We then trained on CUBS for 40 epochs, with a learning rate of 0.01, divided by a factor 10 every 20 epochs. As explained in~\cite{mallya2018packnet}, when pruning the model some parameters of the model are set to zero to make them available for learning subsequent tasks. After pruning, the parameters that remain for the original model need to be finetuned~\cite{mallya2018packnet}. Therefore, postprune finetuning for CUBS was pursued for 10 additional epochs with a learning rate of 0.01. Next, the model was trained on the second task (ImageNet) for 20 epochs, with a learning rate of 0.001, also divided by a factor 10 every ten epochs.
\begin{table}[!h]
    \centering
    \begin{tabular}{l|c |c c}
    \hline
    & & \multicolumn{2}{c}{\# images}\\
    Task & \# classes &  training &  test \\
    \hline
    CUBS& $200$ & $5,994$ & $5,79$4 \\
    ImageNet& $1,000$ & $1,281,144$ & $50,000$ \\ 
    \hline
    \end{tabular}
    \caption{Datasets used for the classification experiments.}
    \label{tab:dataclass}
\end{table}

\paragraph{Results.} Results are reported in Table~\ref{tab:rescls}. We can see that a compression rate of 50\% leads to an accuracy of 64.14\% over the ImageNet test set. Using our proposed adaptive approach, we found that the same ResNet50 network, originally trained with CUBS, could be pruned to a much higher rate, while still keeping the drop of accuracy on CUBS dataset within the 2\% range, therefore saving more capacity for the second task and leading to higher accuracies on ImageNet: 66.98\%. This implies that setting a hard pruning parameter is not optimal in the case of lifelong learning with tasks of different complexity and that learning such rates can make the difference in saving capacity for future tasks. 

This allows us to smartly utilize the available parameters in the model according to the need of the task at hand and still perform within an acceptable loss in performance in the first task, contrarily to classical fine-tuning, where we observed a performance loss of 76.48\% on CUBS (from 77\% of the original model to 0.52\% after finetuning).

 \begin{table}[!h]
 \begin{center}
 \begin{tabular}{l|c|c|c}
 \hline
 & CNN com-& \multicolumn{2}{c}{Accuracy}  \\
Lifelong learning strategy&pression rate  & CUBS  & ImageNet  \\
& (\%) &(\%)& (\%) \\
 \hline

None &  0 & 77.0 & - \\ 
Finetuning ImageNet from CUBS & 0& 0.52 & 67.27 \\
 \hline

PackNet~\cite{mallya2018packnet} &50 & 76.72 & 64.14  \\ 
 \hline

\textbf{AcLL (us)} & 86 & 75.18 & 66.98\\ 
 \hline
 \end{tabular}
 \end{center}
 \caption{Lifelong learning results in the classification setting where a model learned on CUBS is re-used to learn ImageNet as a second task.} \label{tab:rescls}
 \end{table}

\subsection{Semantic segmentation of satellite images}\label{sec:ss}
In this experiment, we aim at learning a sequence of three models dedicated to three different tasks of semantic segmentation of satellite imagery: detecting roads, detecting buildings and mapping coarse landcover types. We investigate if allocating network capacity according to task complexity, for more than two tasks, has an impact on the overall performance. 

\paragraph{Data and setup.} We used the training portion of the DeepGlobe 2018 dataset~\cite{demir2018deepglobe}, which is composed of the disclosed labeled images of the DeepGlobe challenge (\url{deepglobe.org}), as the validation and test sets were unavailable during the course of the challenge. We then divided the data into our own training, validation, and test subsets. We considered three semantic segmentation tasks: `Landcover' (multi-class), `Roads' and `Buildings', the latter two being binary class problems (e.g. road vs. background in the case of `Roads'). In each task, the expected outcome is a map per image, where every pixel is classified in one of the classes or background. The number of images available per task is provided in Table~\ref{tab:data_seg}. 

\begin{table}[!h]
    \centering
    \renewcommand{\tabcolsep}{3pt}
    \begin{tabular}{l|c |c c c |c c}
    \hline
      & & \multicolumn{3}{c|}{\# images} & &\\
      Task & \# classes & training & val & test & { Size (pixels)} & {Resolution (cm)}\\
          \hline
        Roads & 2 & $3,984$ & $1,121$ & $1,121$ & { $1,024\times1,024$} & {50}\\
        Landcover & 7 & $562$ & $120$ & $121$ & { $2,448\times2,448$} & {50}\\ 
        Buildings & 2 & $3,207$ & $687$ & $688$ & { $650\times650$} & {31}\\ 
         \hline
    \end{tabular}
    \caption{Datasets used for the segmentation experiments. Examples of images can be seen in Figures~\ref{fig:segmaps_lrb} and~\ref{fig:segmaps_rlb}. Each image corresponds to a full semantic segmentation map with $r \times c$ pixels to be classified.}
    \label{tab:data_seg}
\end{table}

As base semantic segmentation model, we used ERFNet~\cite{romera2018erfnet}, and evaluated two task sequences: `Landcover', `Road', `Buildings' (\texttt{1:L$\to$R$\to$B}) and `Road', `Landcover' and then `Buildings' (\texttt{2:R$\to$L$\to$B}), respectively. We compared our proposed AcLL against four baselines: finetuning one model after the other, learning without forgetting (LwF~\cite{li2018learning}), an autoencoder-based LwF (AE~\cite{rannen2017encoder}) and fixed-rate compression (PackNet~\cite{mallya2018packnet}).

For all models, we used Adam optimizer~\cite{kingma2015adam} with weight decay of 0.0001. The models were trained for $T = 100$ epochs with an initial learning rate of 0.0005, which was then decreased by a factor of $\left(  1-\frac{t}{T} \right)^{0.9}$ at each epoch $t$~\cite{romera2018erfnet}. All the images were resized to $512 \times 512$ pixels before data augmentation (random horizontal flips and translation of up to two pixels in the horizontal and vertical directions). We used class weight inversely proportional to the number of pixels per class.

The distillation loss weight was set to $1$ and the weight of the autoencoder-based loss component to 0.01 as in~\cite{rannen2017encoder}. Since the task is semantic segmentation, we used a convolutional autoencoder for AE to output a grid of predictions from the bottleneck of ERFNET.

For AcLL, the multitask scheduling was as follows: ERFNet was first trained for $100$ epochs with the first task, and then pruned. After pruning, the model was further fine-tuned on the first task for $30$ epochs. The same scheduling was adopted for the second task, while for the third (the last) task only training with $100$ epochs were performed. For each intermediate task, the drop in accuracy (within 2.0\% from $\mathcal{R}(f_{\theta})$, see Eq.~\eqref{eq:ConstrainedCompression}) was checked on the validation data. The accuracies reported in Tables~\ref{tab:lc_road_build} and~\ref{tab:road_lc_build} are intersection over Union (IoU) scores, evaluated on the test sets of each task.

\paragraph{Results.}
Table~\ref{tab:lc_road_build} presents the results for the first task sequence \texttt{1:L$\to$R$\to$B}, while Table~\ref{tab:road_lc_build} focuses on the second task sequence \texttt{2:R$\to$L$\to$B}. The accuracies for the three individual models trained with just one task are found in the first three rows of both Tables~\ref{tab:lc_road_build} and~\ref{tab:road_lc_build}. We cannot make a direct comparison to the performance reported in the official competition as we did not have access to the official validation and test sets, but we note that the accuracy achieved by our model on our test set is comparable or exceeds that of the accuracies reported by the respective leaderboard winners of the challenge: Landcover $52.24\%$ mIoU \cite{tian2018dense}; Roads $64.12\%$ IoU \cite{zhou2018d}; and Buildings $74.67$ F1-score \cite{hamaguchi2018building}. This is a good indication that we are analyzing the lifelong learning framework on a strong baseline.

 \begin{table}[!t]
 \begin{center}
 \begin{tabular}{p{3.5cm}|c|c|c|c || c}
 \toprule
  Lifelong & CNN com- & \multicolumn{3}{c||}{Accuracy (\%)} & 3 tasks  \\
 learning&pression rate  & Task 1 & Task 2 & Task 3 & average \\
  strategy  & \small{(\%$_{T1,2}$), (\%$_{T2,3}$)}& \small{Landcover} & \small{Roads} & \small{Buildings} & accuracy \\
\toprule 
 \multirow{3}{*}{None (baselines)} &  0 & 48.20 & - & - \\
 &0 & - & 71.03 & - \\
 &0 & - & - & 80.10\\
 \hline
  Fine-tune \{T1, T2\}$\to$T3  &0& 3.15 & 47.95 & 79.26 &43.45 \\
LwF~\cite{li2018learning} & 0& 26.52 & 62.53 & \underline{81.30} & 56.78\\
  AE~\cite{rannen2017encoder} & 0& 25.77 & 64.59 & \textbf{81.36} & 57.24\\
  \midrule\midrule
PackNet~\cite{mallya2018packnet} &(50.0), (50.0) & \textbf{49.36} &  {67.67} & 75.24 & {64.09} \\
 &(75.0), (75.0) & \underline{47.47} & \underline{68.81} & 78.85 & \underline{65.04} \\

 \hline

\textbf{AcLL (us)} & (84.375), (72.0)  & 47.30 &  \textbf{68.92} & 79.14 & \textbf{65.12}  \\
\bottomrule
 \end{tabular}
 \end{center}
 \caption{Sequential learning of tasks: Landcover $\to$ Roads $\to$ Buildings (best result in bold, second best underlined).}\label{tab:lc_road_build}
 \end{table}

\begin{table}[!t]
 \begin{center}
 \begin{tabular}{p{3.5cm}|c|c|c|c || c}
 \toprule
  Lifelong & CNN com- & \multicolumn{3}{c||}{Accuracy (\%)} & 3 tasks  \\
 learning&pression rate  & Task 1 & Task 2 & Task 3 & average \\
  strategy  & \small{(\%$_{T1,2}$), (\%$_{T2,3}$)}& \small{Roads} & \small{Landcover} & \small{Buildings} & accuracy \\
\toprule 
 \multirow{3}{*}{None (baselines)} &  0 & 71.03& - & - \\
 &0 & - & 48.20 & - \\
 &0 & - & - & 80.10\\
 \hline
  Fine-tune \{T1, T2\}$\to$T3  &0& 47.95 & 2.25 & 79.75 & 43.31 \\
LwF~\cite{li2018learning} & 0&  62.71 & 22.70 & \textbf{81.23} & 55.54\\
  AE~\cite{rannen2017encoder} & 0& 62.73 & 30.92 & \underline{80.52} & 58.05\\
  \midrule\midrule
PackNet~\cite{mallya2018packnet} &(50.0), (50.0) & \textbf{70.75} &  \textbf{53.72} & 70.48 & \textbf{64.98}  \\
 &(75.0), (75.0) & \underline{70.22} & \underline{48.80} & 74.61 & 64.54\\

 \hline

\textbf{AcLL (us)}  & (86.25), (92.0)  & {69.08} &  48.13 & 77.53 & \underline{64.91}  \\
\bottomrule
 \end{tabular}
 \end{center}
 \caption{Sequential learning of tasks: Roads $\to$ Landcover $\to$ Buildings (best result in bold, second best underlined).}\label{tab:road_lc_build}
 \end{table}

In both the Tables~\ref{tab:lc_road_build} and~\ref{tab:road_lc_build}, the baseline Fine-tune obtains good results in the last task, \ie the detection of buildings. However, its performance on the other two tasks is very poor. The baselines LwF and AE obtain the best results on the last task. However, they show heavily degraded performances on the first task (a drop of more than 20\%, see Table~\ref{tab:lc_road_build}) and, to a lesser extent, on the second task (drop by 8\% Table~\ref{tab:road_lc_build}). It is evident that these lifelong learning baselines fail in remembering the previous tasks compared with network compression approaches (PackNet and our AcLL), which always outperform the competing methods by a large margin, leading to the best average score over the three tasks (last column of both the Tables). Moreover, our proposed AcLL achieves the best or second-best accuracy in both task orderings (Tables~\ref{tab:lc_road_build} and~\ref{tab:road_lc_build}) and also allows for the compression of the network according to the task's complexity at hand. Such optimal compression is way far greater than 50\%, especially since the tasks are not so complex and an efficient network can be obtained with higher compression rates. This approach allows for freeing more redundant parameters for future, unseen tasks if the current task is small or less complex. 
The benefits of AcLL with respect to PackNet can be seen in the last task, where the additional freeing of parameters provides more capacity, and therefore a more accurate CNN for the third task. In Table~\ref{tab:road_lc_build} PackNet with 50\% compression ratio achieved a marginally higher average accuracy {over the three tasks}, it did so at the cost of a more than 7\% reduction in accuracy on the final task and after exhausting 75\% of its available parameters after two tasks, while the adaptive method had only used less than 21\% of the available weights. We expect the benefits to become more and more evident with increase in the number of tasks.

Inspecting the segmentation maps for the two tasks sequences (Figures~\ref{fig:segmaps_lrb} and~\ref{fig:segmaps_rlb}, respectively), we observe that our adaptive pruning AcLL leads to overall more accurate maps than the three competing baselines, which provide accurate maps mostly for the last task. The segmentation maps for the three tasks (`Landcover', `Roads' and `Buildings') are closer to their respective ground truths (Column 2 of both figures). By looking at the maps, it becomes evident that the three competing methods struggle to remember the correct decision function for the first task and only partially perform adequately on the second. The proposed AcLL provides plausible maps for all tasks, even if it sometimes hallucinates linear structures (though still topologically plausible) for the road task in the \texttt{2:R$\to$L$\to$B} sequence (see second row, column AcLL in Fig~\ref{fig:segmaps_rlb}).

\newcommand{\wdt}{.15}
\begin{figure*}[!t]
\begin{center}
\renewcommand{\tabcolsep}{2pt}
\scriptsize{
\begin{tabular}{ccc|cccc} 

&Image & Ground truth & Finetune & LwF & AE & AcLL \\
  \rotatebox{90}{T1: Landcover} &      
   \includegraphics[width=\wdt\columnwidth]{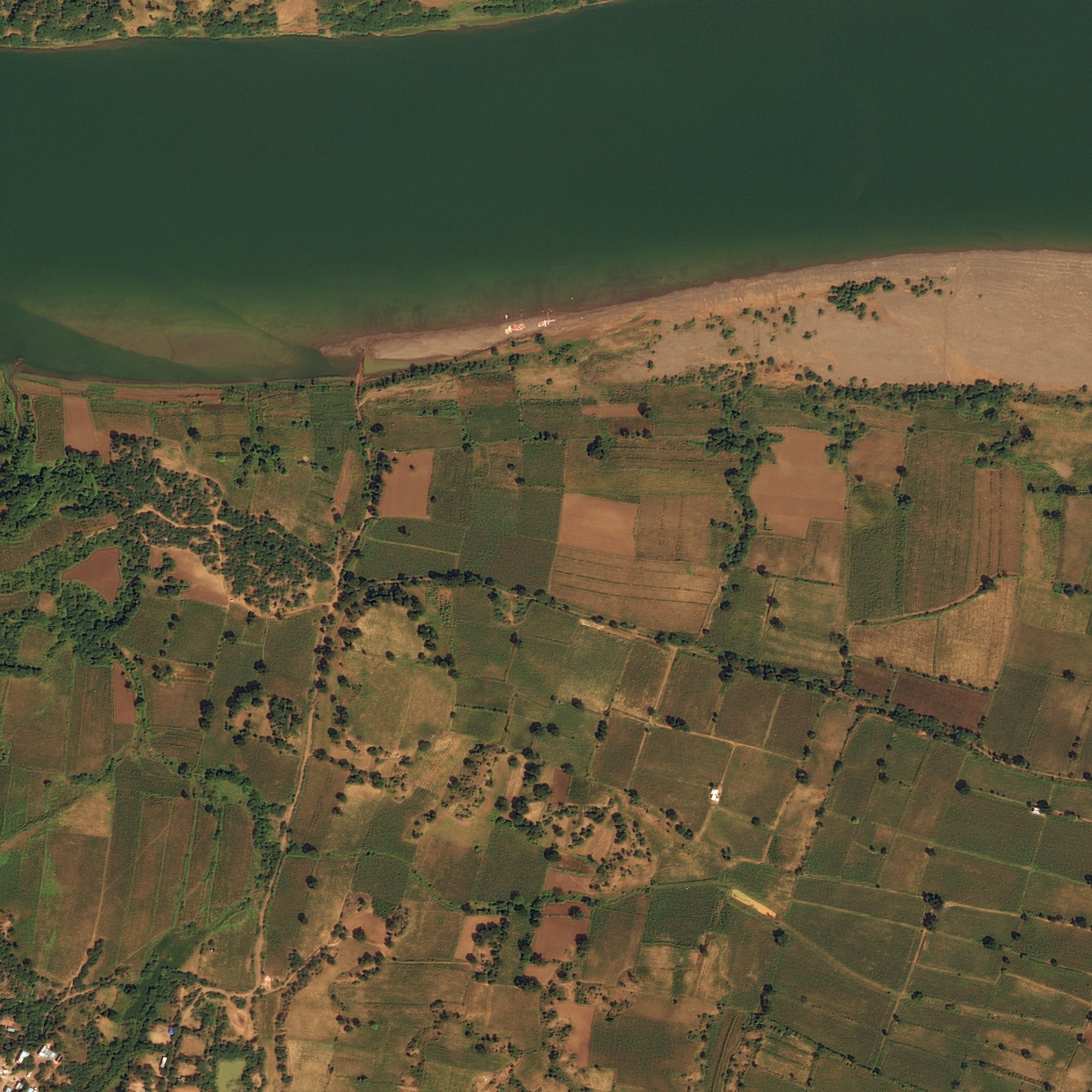} &    
       \includegraphics[width=\wdt\columnwidth]{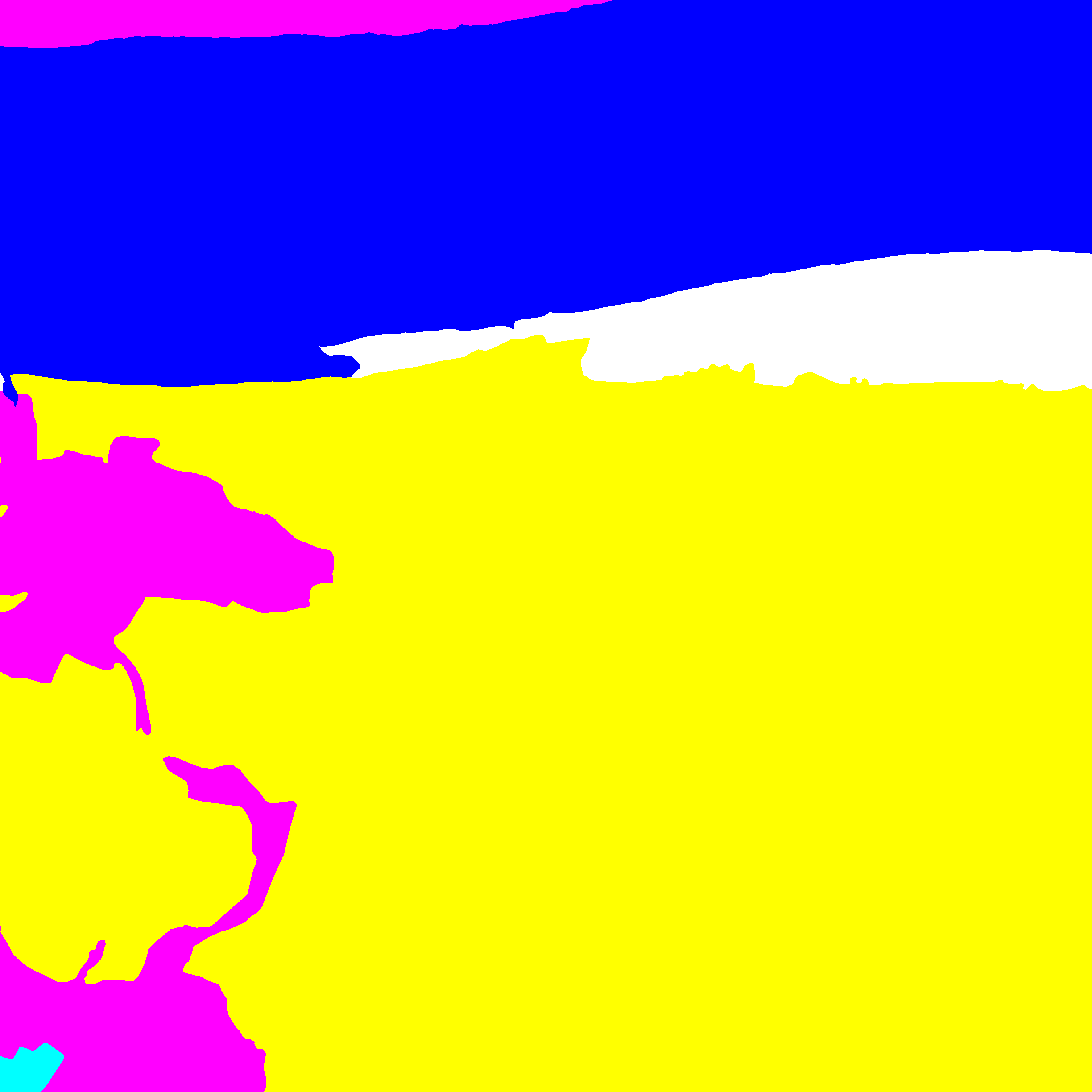} & 
            \includegraphics[width=\wdt\columnwidth]{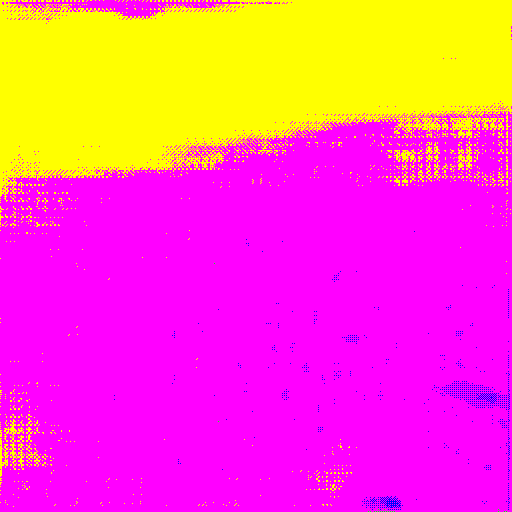} & 
            \includegraphics[width=\wdt\columnwidth]{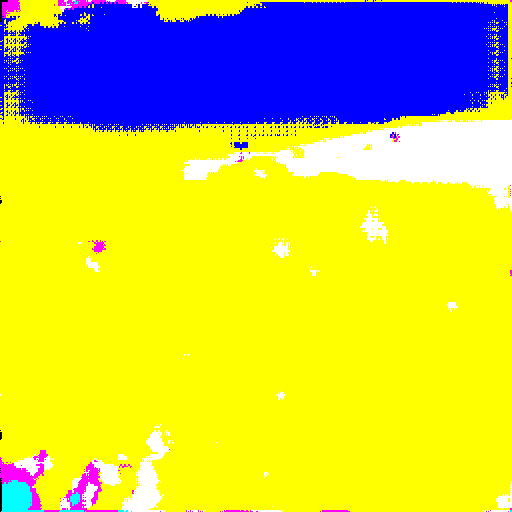} & 
            \includegraphics[width=\wdt\columnwidth]{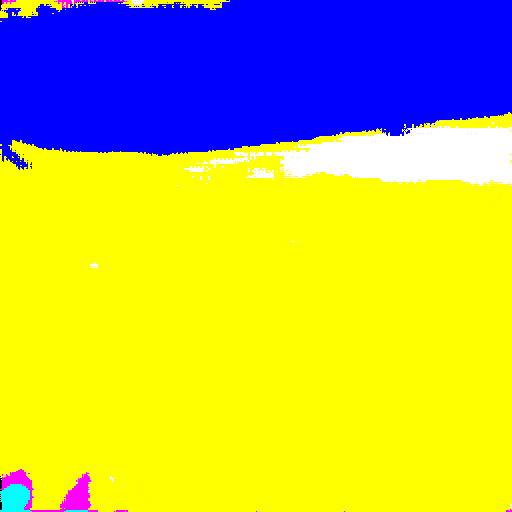} & 
            \includegraphics[width=\wdt\columnwidth]{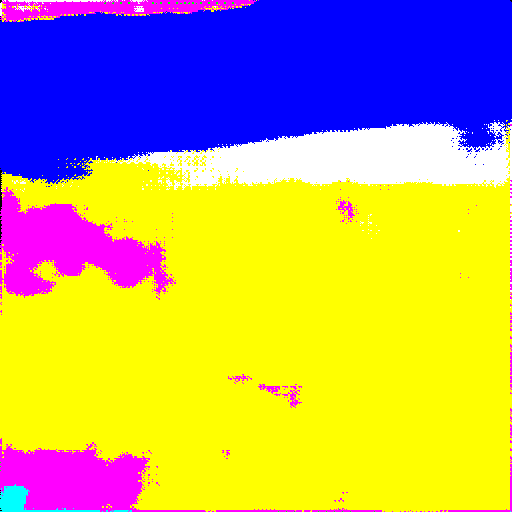} \\
 \rotatebox{90}{T2: Roads} &
 \includegraphics[width=\wdt\columnwidth]{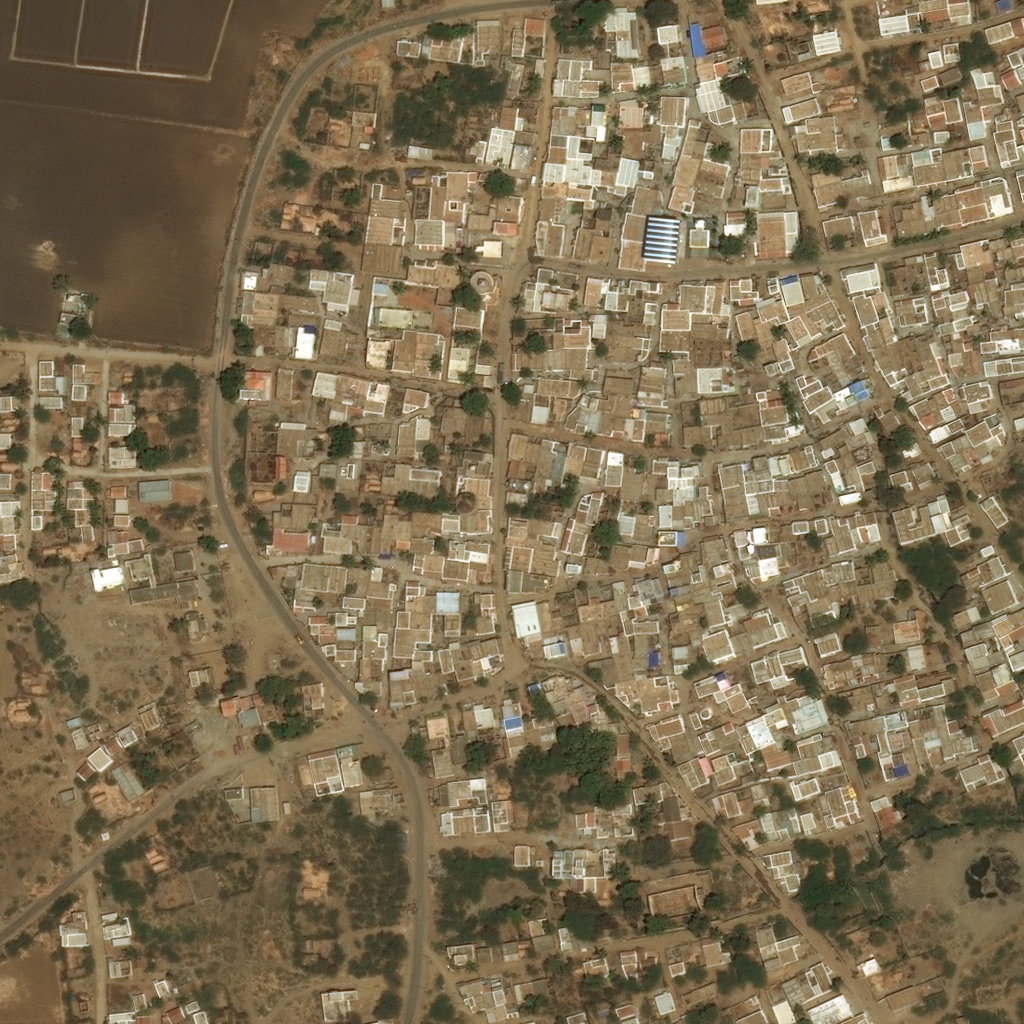} & 
 \includegraphics[width=\wdt\columnwidth]{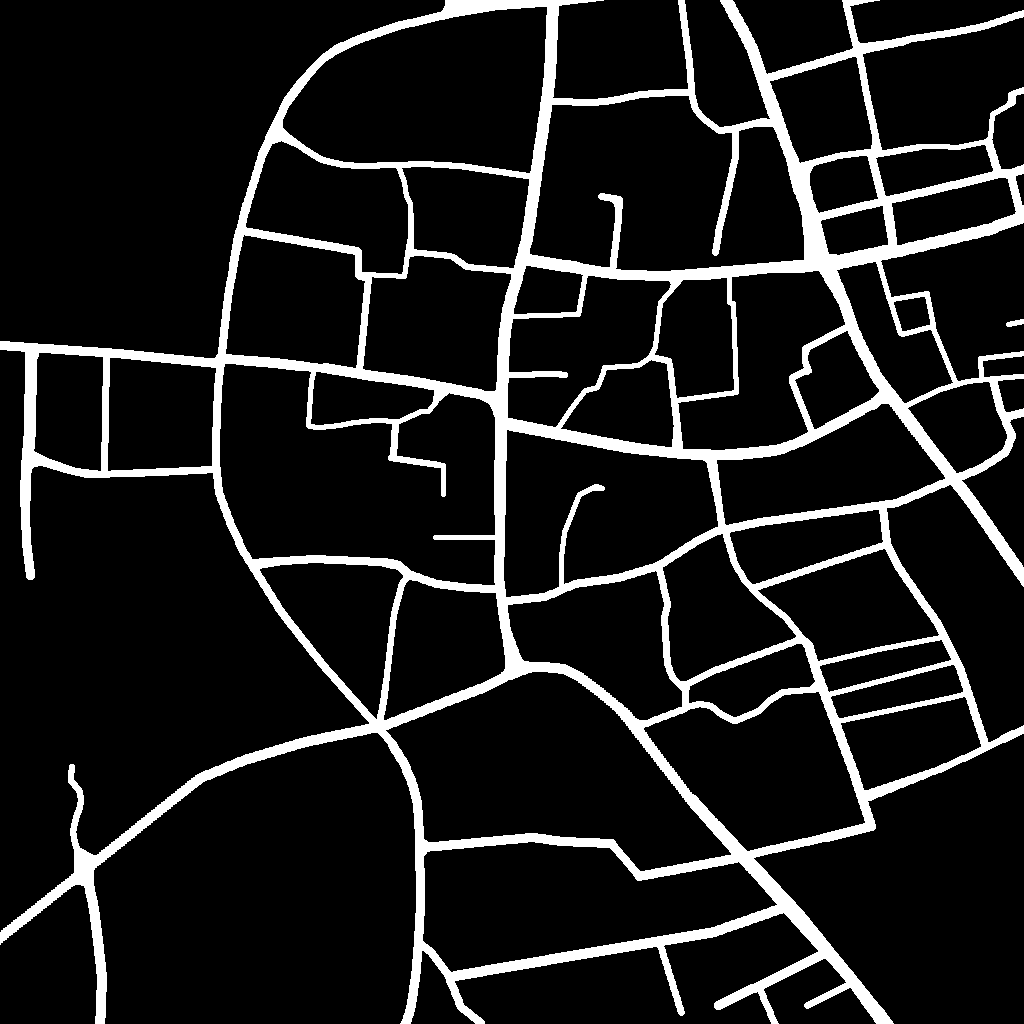} & 
  \includegraphics[width=\wdt\columnwidth]{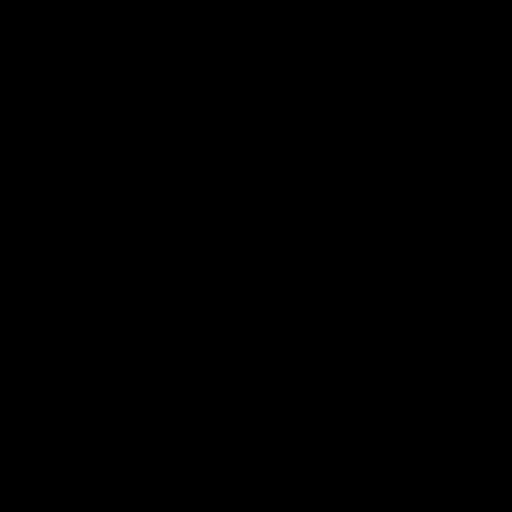} & 
      \includegraphics[width=\wdt\columnwidth]{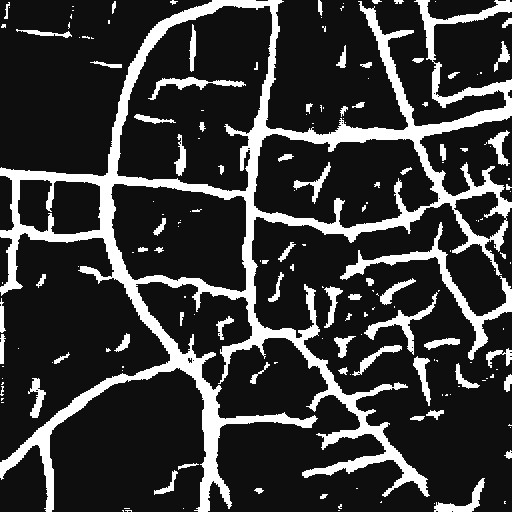} & 
        \includegraphics[width=\wdt\columnwidth]{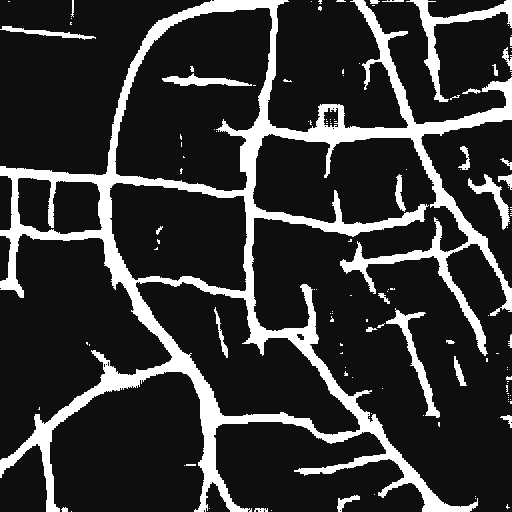} & 
          \includegraphics[width=\wdt\columnwidth]{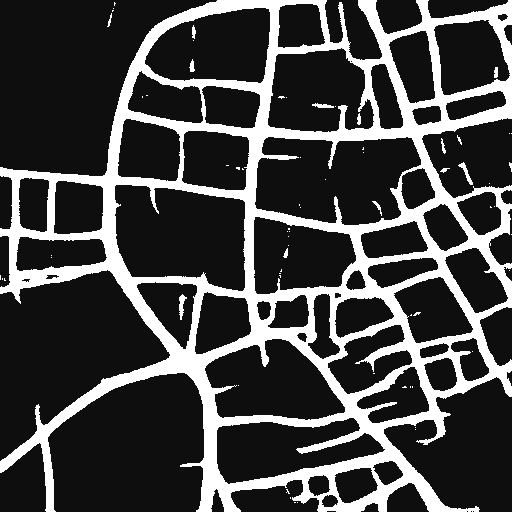} \\
   \rotatebox{90}{T3: Buildings} &                 
              \includegraphics[width=\wdt\columnwidth]{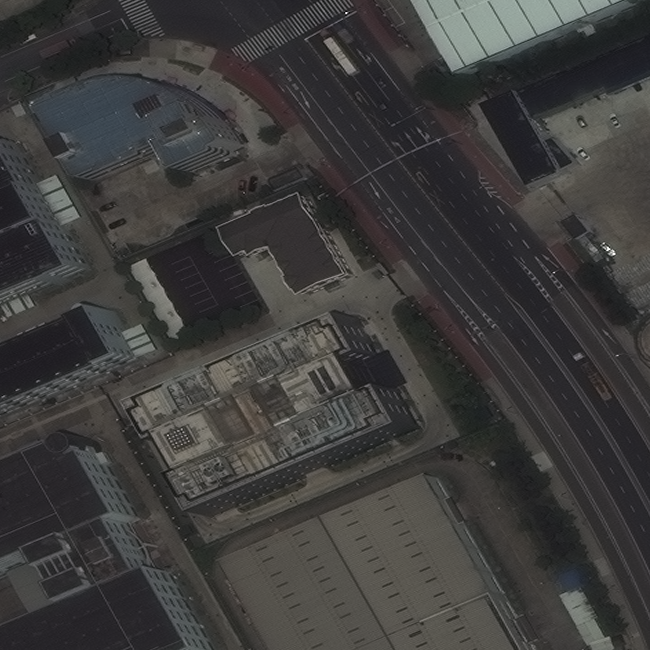}&
                \includegraphics[width=\wdt\columnwidth]{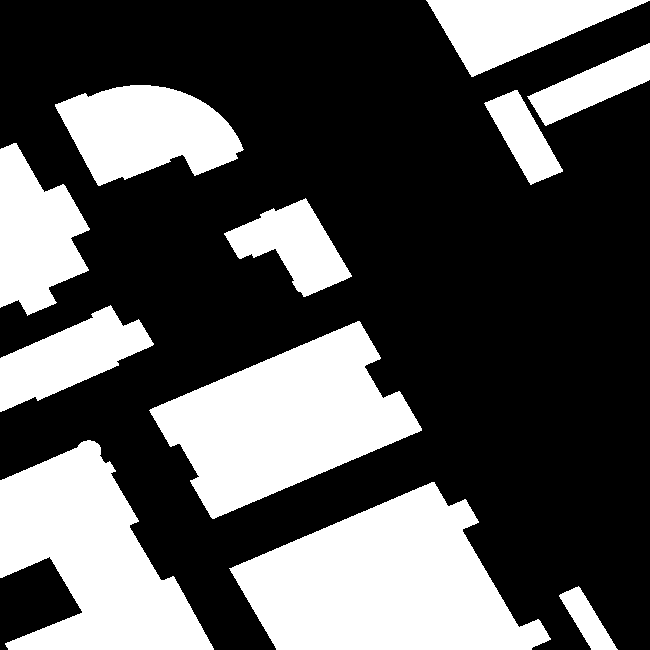}&
                \includegraphics[width=\wdt\columnwidth]{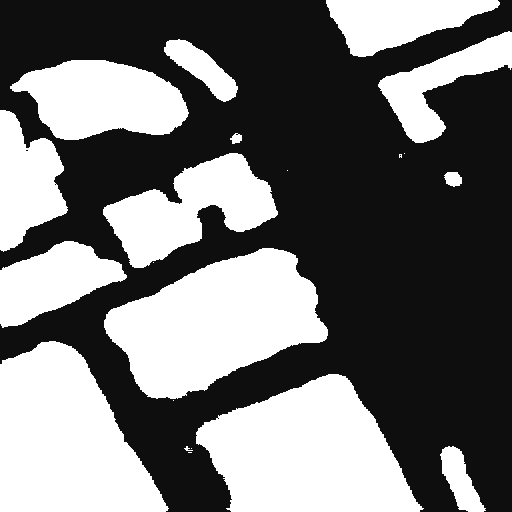} &
                \includegraphics[width=\wdt\columnwidth]{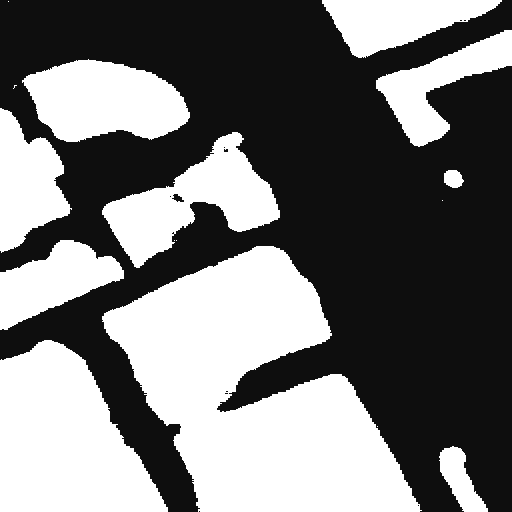} &
                \includegraphics[width=\wdt\columnwidth]{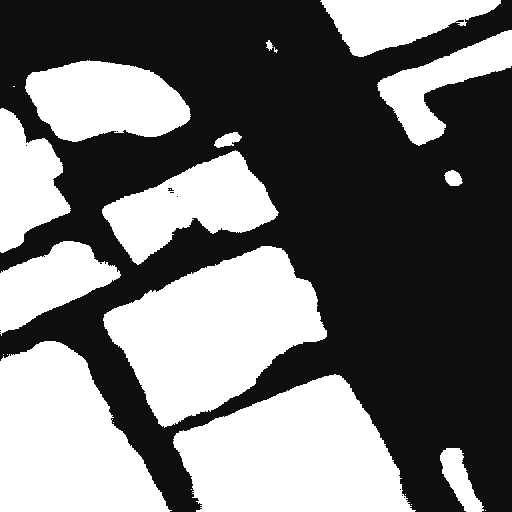} &
                \includegraphics[width=\wdt\columnwidth]{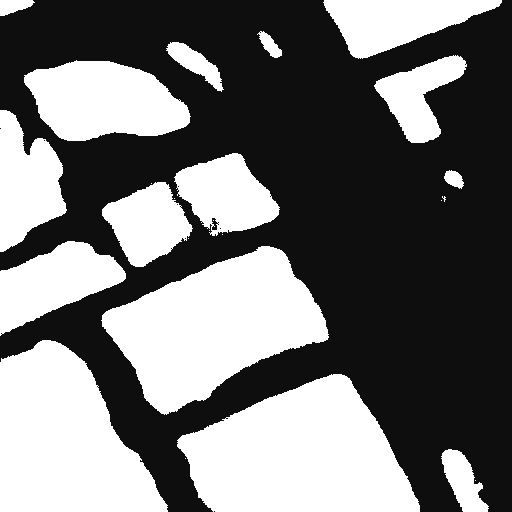} \\
\end{tabular} 
}
\end{center}
\caption{Comparison of the different lifelong learning strategies in the DeepGlobe data for the sequence: Landcover $\to$ Roads $\to$ Buildings (for AcLL: 84.375\%, 72\%). The legend for the Landcover task is: {\color{green} \textbf{forest}}, {\color{blue} \textbf{water}}, {\color{cyan} \textbf{urban}}, {\color{magenta} \textbf{rangeland}}, {\color{yellow} \textbf{agriculture}}; white shows barren land and black denotes background/unknown.}
\label{fig:segmaps_lrb}
\end{figure*}

\begin{figure*}[!t]
\begin{center}
\renewcommand{\tabcolsep}{2pt}
\scriptsize{
\begin{tabular}{ccc|cccc} 
&Image & Ground truth & Finetune & LwF & AE & AcLL \\
\rotatebox{90}{T1: Roads} &
 \includegraphics[width=\wdt\columnwidth]{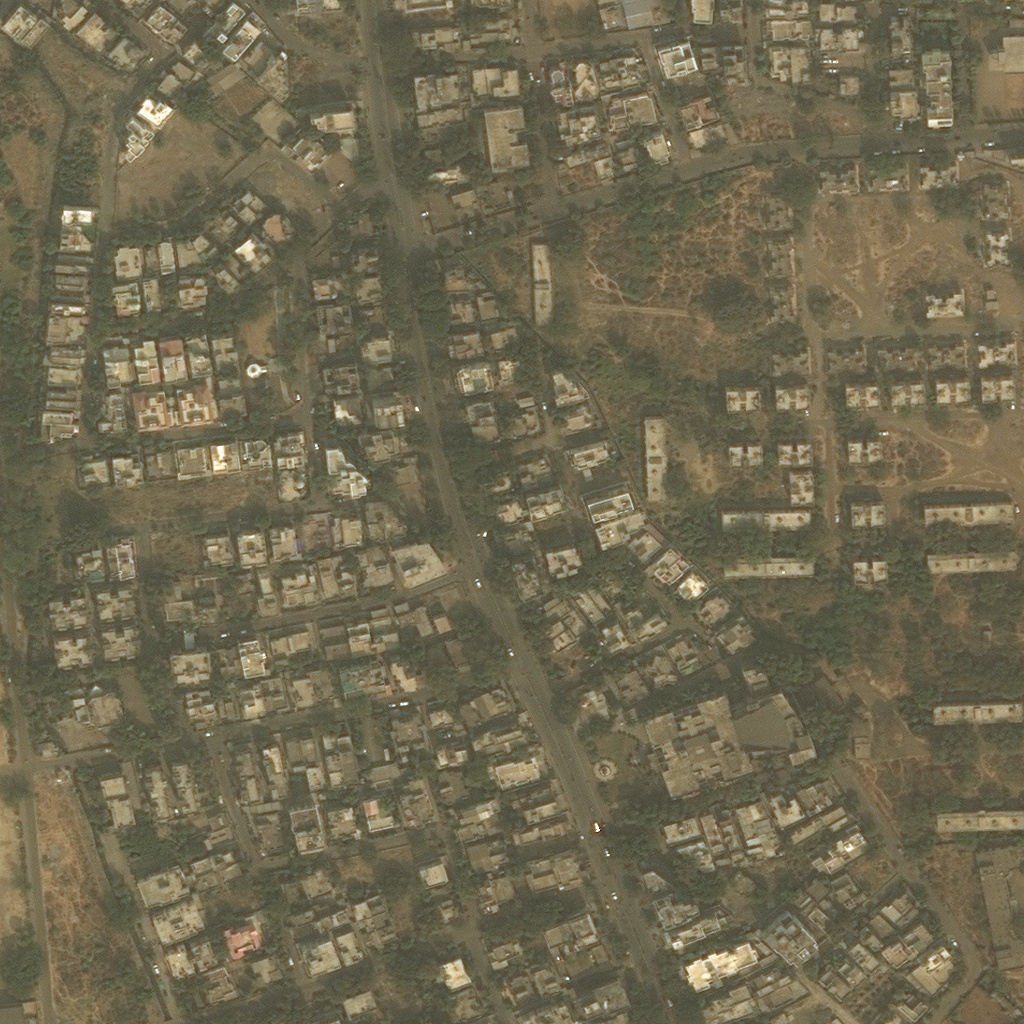} & 
 \includegraphics[width=\wdt\columnwidth]{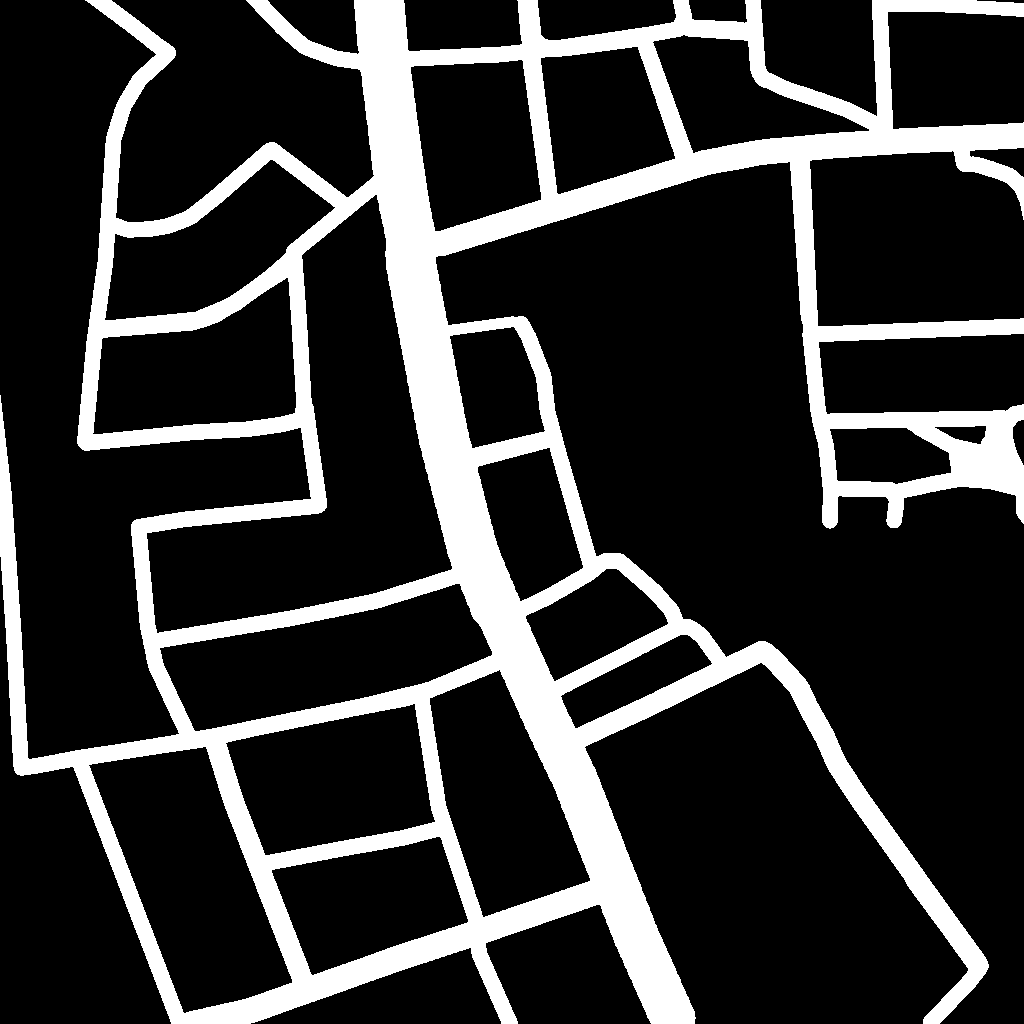} & 
  \includegraphics[width=\wdt\columnwidth]{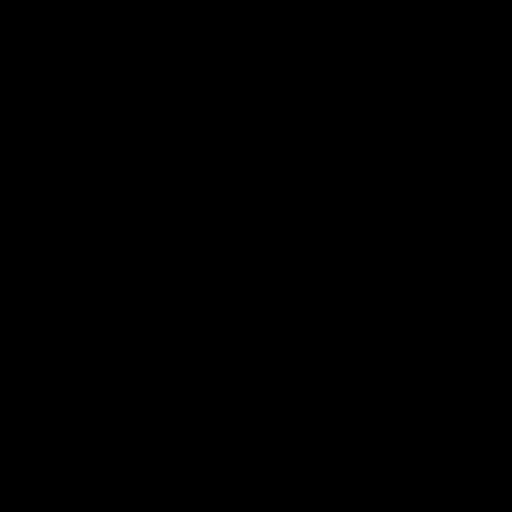} & 
      \includegraphics[width=\wdt\columnwidth]{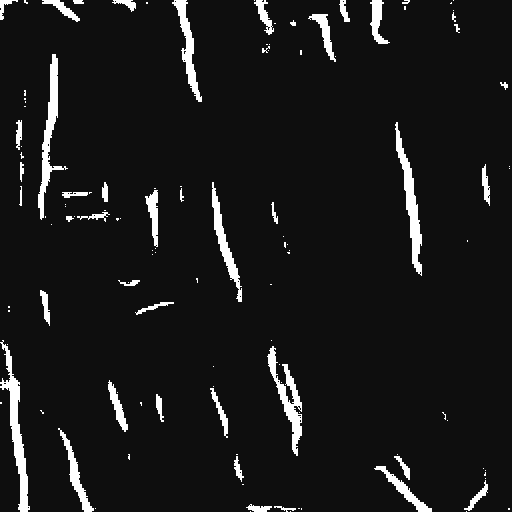} & 
        \includegraphics[width=\wdt\columnwidth]{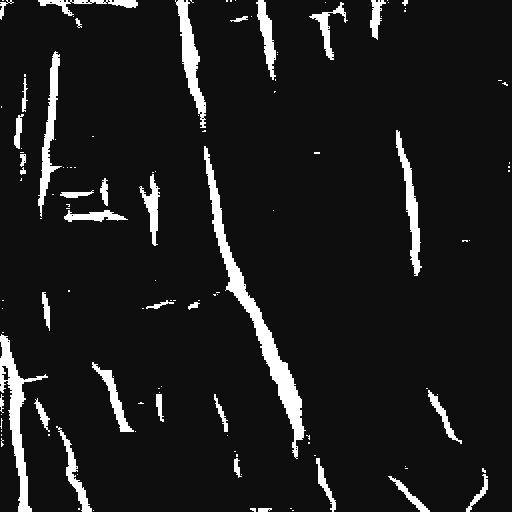} & 
          \includegraphics[width=\wdt\columnwidth]{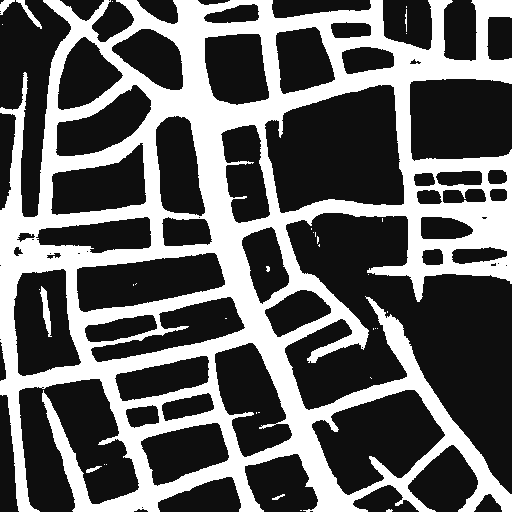} \\
 \rotatebox{90}{T2: Landcover} &      
   \includegraphics[width=\wdt\columnwidth]{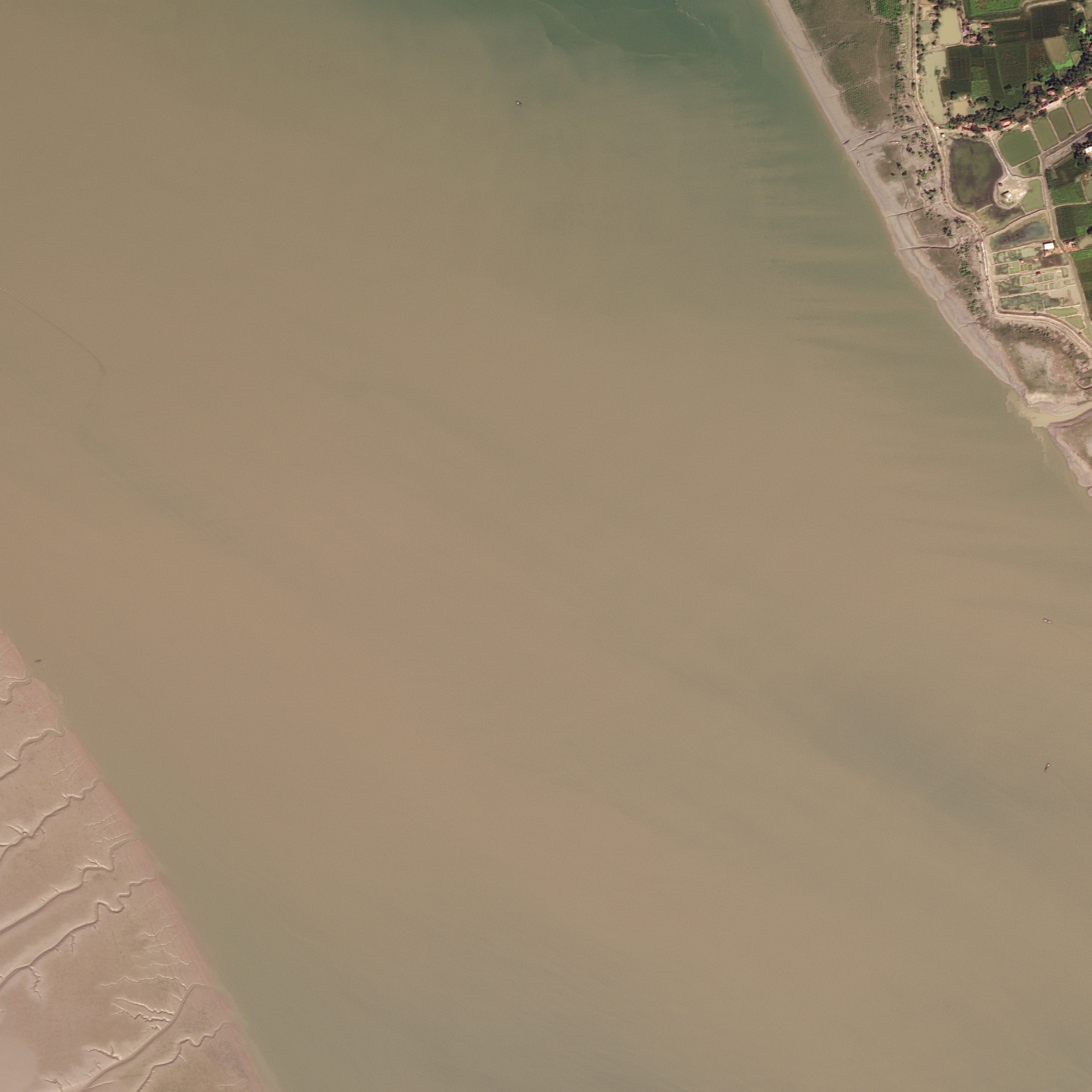} &    
       \includegraphics[width=\wdt\columnwidth]{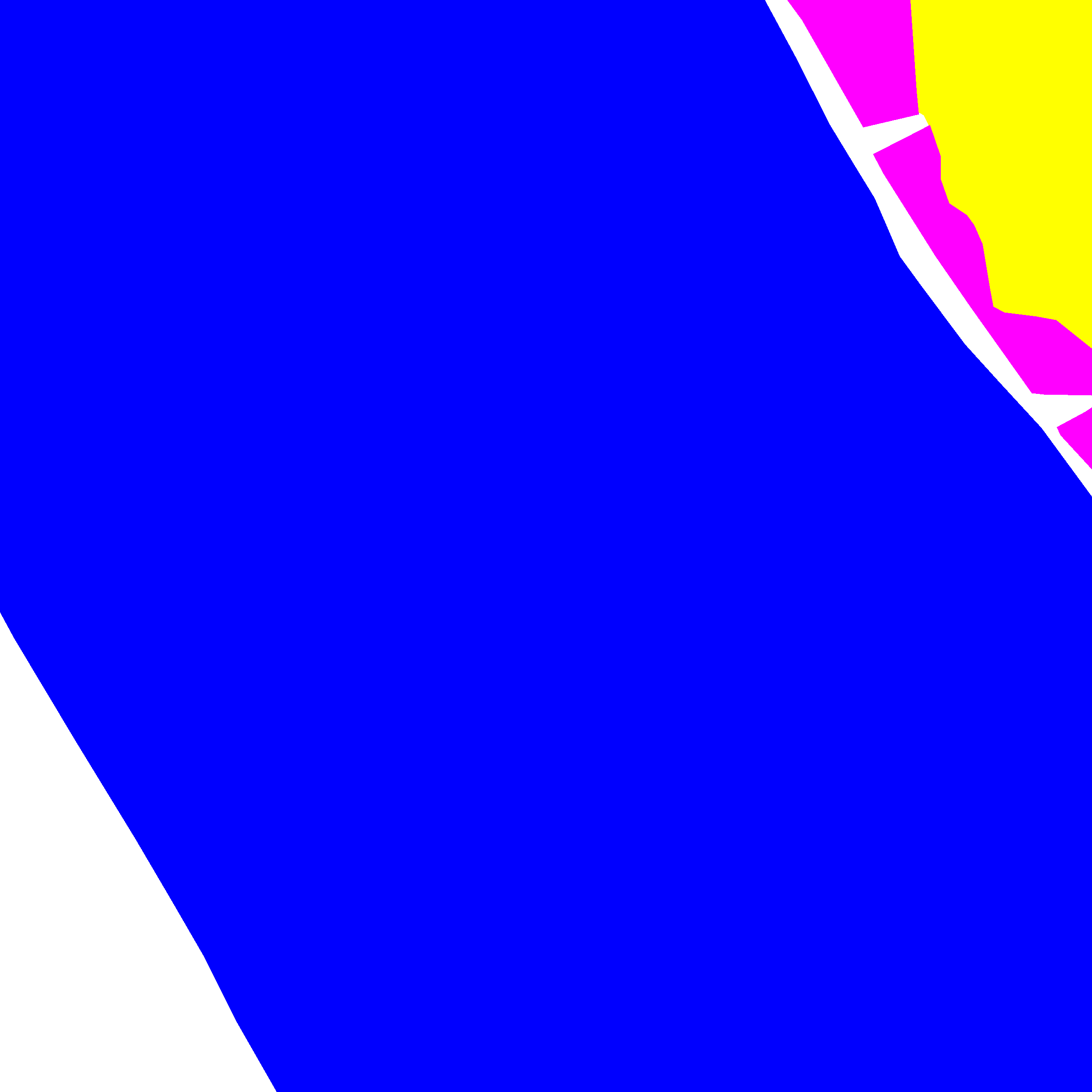} & 
            \includegraphics[width=\wdt\columnwidth]{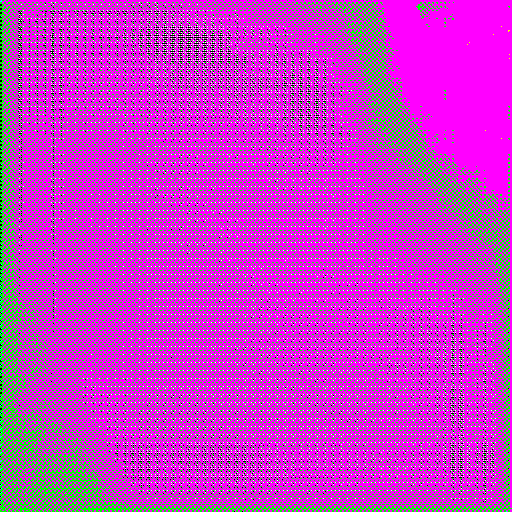} & 
            \includegraphics[width=\wdt\columnwidth]{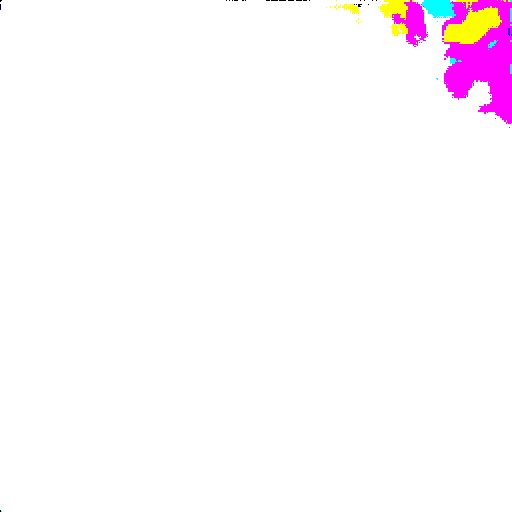} & 
            \includegraphics[width=\wdt\columnwidth]{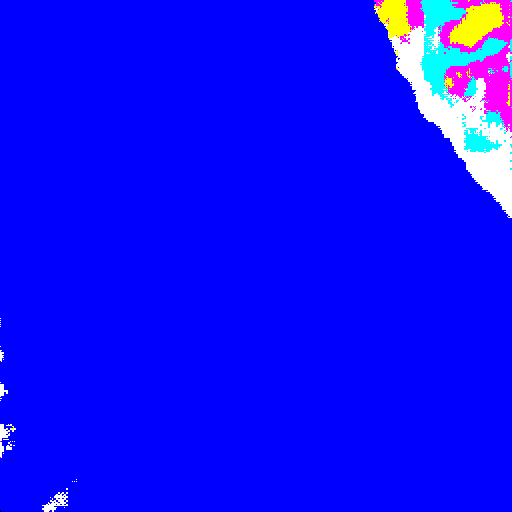} & 
            \includegraphics[width=\wdt\columnwidth]{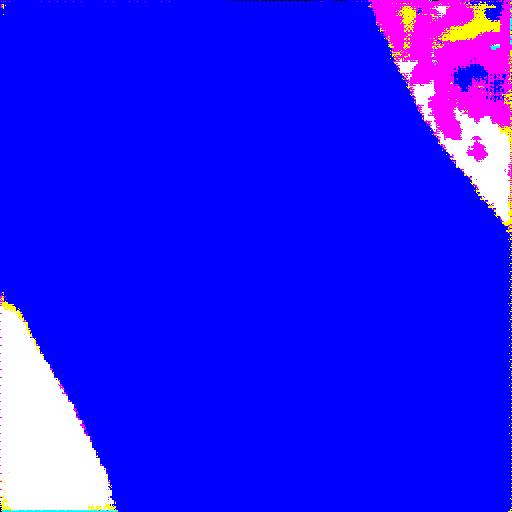} \\
   \rotatebox{90}{T3: Buildings} &                 
              \includegraphics[width=\wdt\columnwidth]{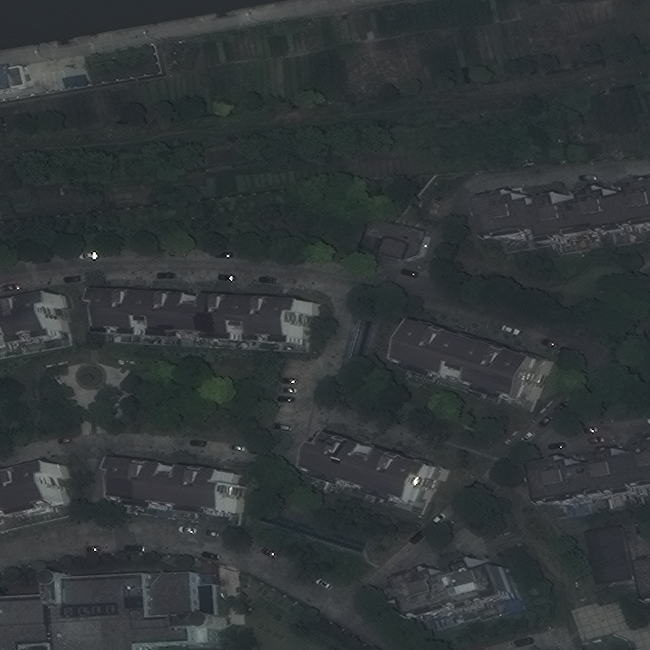}&
                \includegraphics[width=\wdt\columnwidth]{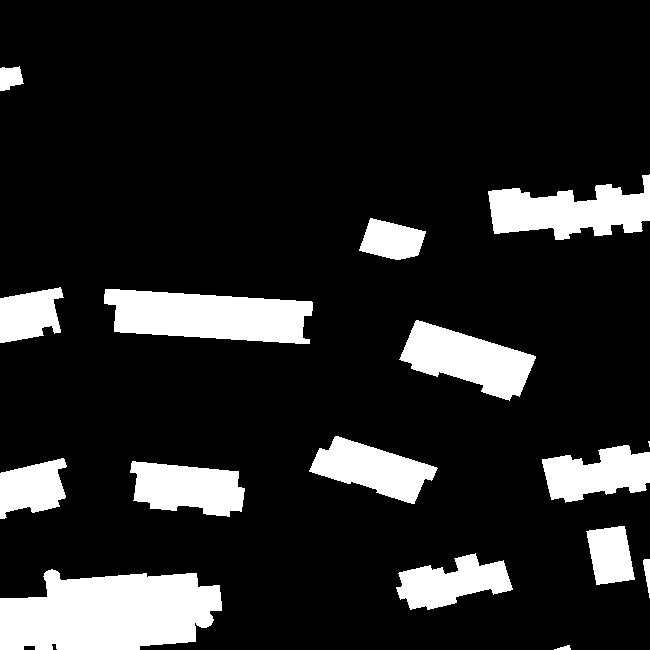}&
                \includegraphics[width=\wdt\columnwidth]{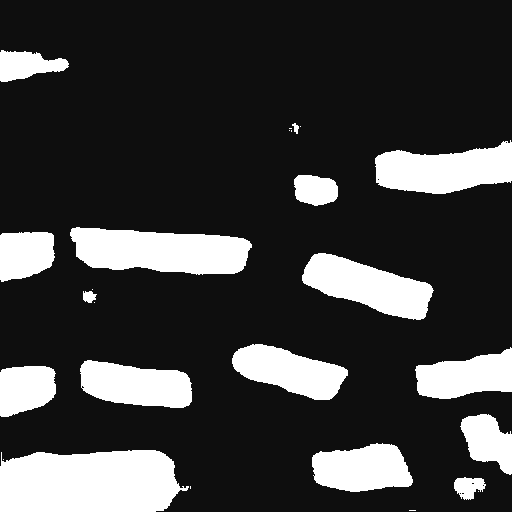} &
                \includegraphics[width=\wdt\columnwidth]{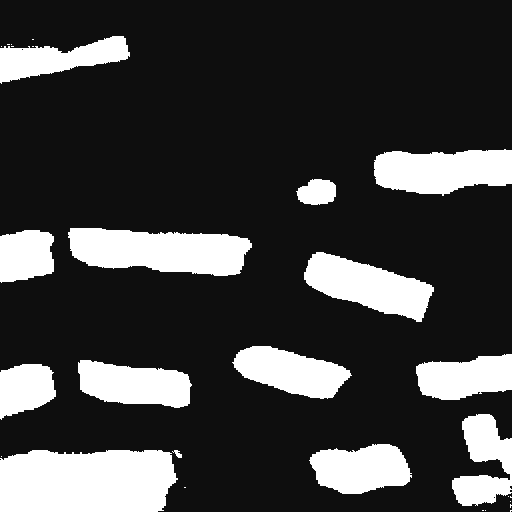} &
                \includegraphics[width=\wdt\columnwidth]{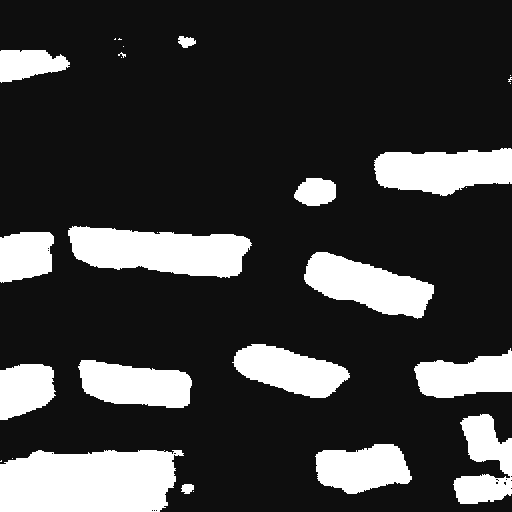} &
                \includegraphics[width=\wdt\columnwidth]{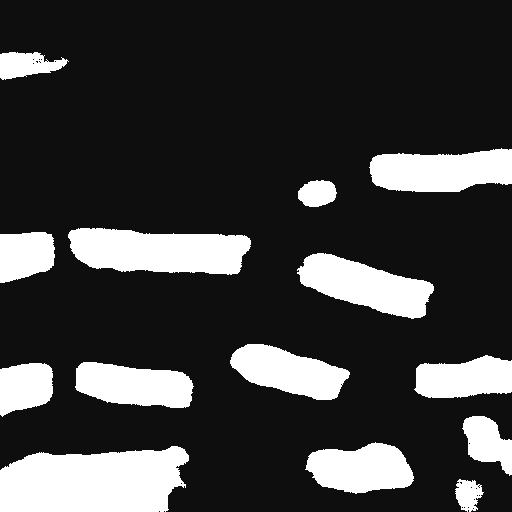} \\
\end{tabular} 
}
\end{center}
\caption{Comparison of the different lifelong learning strategies in the DeepGlobe data for the sequence: Roads $\to$ Landcover $\to$ Buildings (for AcLL: 86.25\%, 92\%). The legend for the Landcover task is: {\color{green} \textbf{forest}}, {\color{blue} \textbf{water}}, {\color{cyan} \textbf{urban}}, {\color{magenta} \textbf{rangeland}}, {\color{yellow} \textbf{agriculture}}; white shows barren land and black denotes background/unknown.}
\label{fig:segmaps_rlb}
\end{figure*}

\section{Conclusion}\label{sec:discussion_and_conclusion}
In this work, we propose a method for lifelong learning based on adaptive compression. Different from a recent compression-based method, called PackNet~\cite{mallya2018packnet}, that uses a pre-defined compression rate, we perform adaptive compression that considers the complexity of the task at hand while maintaining guarantees on accuracies of the compressed network on previous tasks. Thus, if a model was trained for a relatively simple task it can be strongly compressed in a way that more free parameters are available to train other subsequent tasks. Our experimental results show the advantage of our AcLL method over four baseline methods: standard finetuning, Learning without Forgetting (LwF), Encoder-based life long learning (AE), and PackNet.
\section*{Acknowledgement}
S.S. is thankful to Gabriel Peyr\'{e} (``ERC project Noria") and Marco Cuturi (``Chaire d'excellence de l'IDEX Paris Saclay") for their support by providing access to computing facilities. M.B.\ and M.B.B.\ acknowledge support from FWO (grant G0A2716N), an Amazon Research Award, an NVIDIA GPU grant, and the Facebook AI Research Partnership.  This research was carried out while S.S.\ was a visiting researcher at KU Leuven.

\bibliography{egbib}

\end{document}